\documentclass[10pt,twocolumn,letterpaper]{article}

\usepackage[pagenumbers]{cvpr} % To force page numbers, e.g. for an arXiv version

\usepackage[dvipsnames]{xcolor}
\usepackage{color}

\usepackage{amssymb}% http://ctan.org/pkg/amssymb
\usepackage{pifont}% http://ctan.org/pkg/pifont
\newcommand{\cmark}{\ding{51}}%
\newcommand{\xmark}{\ding{55}}%

\usepackage{algorithm2e}
\RestyleAlgo{ruled}
\newlength\mylen
\newcommand\ExtIn[1]{%
  \settowidth\mylen{\KwIn{}}%
  \setlength\hangindent{\mylen}%
  \hspace*{\mylen}#1\\}

\newcommand\blfootnote[1]{% 
\begingroup 
\renewcommand\thefootnote{}\footnote{#1}% 
\addtocounter{footnote}{-1}% 
\endgroup 
}

\definecolor{cvprblue}{rgb}{0.21,0.49,0.74}
\usepackage[pagebackref,breaklinks,colorlinks,citecolor=cvprblue]{hyperref}
\usepackage[accsupp]{axessibility} % Improves PDF readability for those with visual impairments.

\def\paperID{12208} % *** Enter the Paper ID here
\def\confName{CVPR}
\def\confYear{2024}

\title{Neural 3D Strokes: Creating Stylized 3D Scenes with Vectorized 3D Strokes}

\author{Hao-Bin Duan\textsuperscript{1} \quad Miao Wang\textsuperscript{1,2\text{\textbf{*}}} \quad Yan-Xun Li\textsuperscript{1} \quad Yong-Liang Yang\textsuperscript{3}\\
\textsuperscript{1}State Key Laboratory of Virtual Reality Technology and Systems, SCSE, Beihang University \\
\textsuperscript{2}Zhongguanchun Laboratory  \quad   \textsuperscript{3}Department of Computer Science, University of Bath \\
}

\begin{document}
%\begin{bibunit}

%=========================================================================

\twocolumn[{%
 \renewcommand\twocolumn[1][]{#1}%
 \maketitle
 \begin{center}
   \centering
   \includegraphics[width=\textwidth]{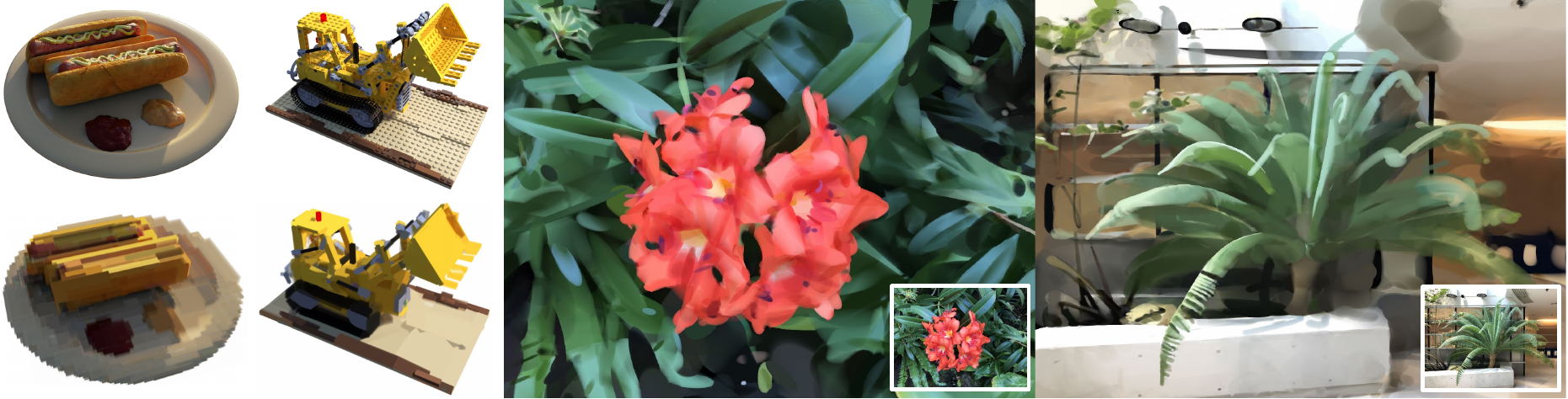}
   \captionof{figure}{We propose a method to stylize a 3D scene from multi-view 2D images using vectorized 3D strokes based on geometric primitives and splines. The four scenes from left to right are drawn with axis-aligned box, oriented box, ellipsoid, and cubic Bézier curve, respectively. }
   \label{fig:teaser}
 \end{center}
 }]

%=========================================================================

\begin{abstract}
\vspace{-2.0mm}

We present Neural 3D Strokes, a novel technique to generate stylized images of a 3D scene at arbitrary novel views from multi-view 2D images. Different from existing methods which apply stylization to trained neural radiance fields at the voxel level, our approach draws inspiration from image-to-painting methods, simulating the progressive painting process of human artwork with vector strokes. We develop a palette of stylized 3D strokes from basic primitives and splines, and consider the 3D scene stylization task as a multi-view reconstruction process based on these 3D stroke primitives. Instead of directly searching for the parameters of these 3D strokes, which would be too costly, we introduce a differentiable renderer that allows optimizing stroke parameters using gradient descent, and propose a training scheme to alleviate the vanishing gradient issue. The extensive evaluation demonstrates that our approach effectively synthesizes 3D scenes with significant geometric and aesthetic stylization while maintaining a consistent appearance across different views. Our method can be further integrated with style loss and image-text contrastive models to extend its applications, including color transfer and text-driven 3D scene drawing. 
Results and code are available at \url{http://buaavrcg.github.io/Neural3DStrokes}.

\end{abstract}
\vspace{-4.0mm}

\blfootnote{\text{\textbf{*}} Corresponding author.}

%=========================================================================

\section{Introduction}
\label{sec:intro}

Artistic style image creation, historically a domain requiring significant skill and time, has been revolutionized by neural network-based Style Transfer techniques~\cite{gatys2016image,gatys2015neural}. These methods usually separate, manipulate, and merge the content and style of images to create an artistic effect. Extending this to 3D scenes, however, is challenging due to complex geometries and appearance traits~\cite{shen2018neural}, and traditional convolutional neural network (CNN) methods~\cite{risser2017stable} are not readily adaptable to 3D spaces. Recently, 3D implicit scene representations, particularly Neural Radiance Fields (NeRF), have become popular, since they are fully differentiable and easy to optimize. Efforts to apply artistic styles to NeRF~\cite{liu2023stylerf,nguyen2022snerf,Huang22StylizedNeRF} often involve separate NeRF training and style transfer, followed by integration style transfer loss from multiple views~\cite{fan2022unified,zhang2022arf,chiang2022stylizing}. The fundamental approach continues to be the independent manipulation of content and style. Nevertheless, existing NeRF stylization techniques only alter the color aspects, leaving the density-and consequently the geometry—unchanged, leading to stylizations that lack geometric changes.

Style transfer methods in 2D and 3D primarily manipulate pixels or voxels to achieve artistic effects, differing from traditional art where artists use brushstrokes and a degree of randomness in brushes, materials, and colors. This traditional approach is time-consuming and skill-intensive. To automate it, research has explored image generation with vectorized strokes~\cite{hertzmann1998painterly}, defined by position, color, size, and direction, and optimized to match target images. 3D painting tools like Google's Tilt Brush~\cite{chittenden2018tilt} use VR/AR to create art scenes with various brushes in virtual spaces. However, the field of automated stroke-based 3D art generation remains relatively unexplored.

In this work, we present a novel technique for transforming 2D images with known poses into stylized 3D scenes using vectorized 3D strokes. Our approach recreates 3D scenes that exhibit distinct geometric and appearance styles, emulating the stroke-by-stroke painting process employed by human artists. The vector-based nature of our system enables rendering at any desired resolution. It accommodates a range of 3D stroke styles and supports more intricate stylizations and generations through the use of style and semantic losses. The core of our method is a differentiable, vectorized 3D scene representation, which diverges from NeRF by using parameterized 3D strokes to synthesize 2D images at different viewpoints. This differentiable representation allows for the direct optimization of stroke parameters via gradient descent, sidestepping the need for previous greedy-search-based or reinforcement learning-based prediction methods in stroke-based image synthesis. Furthermore, we analyze the gradient behaviors of our stroke-based 3D representation and introduce a stroke initialization and update scheme that avoids sub-optimal initialization where optimization is difficult due to local minima.

Our method was evaluated using the multi-view datasets of real-world and synthetic images. Our experiments demonstrate that it effectively creates high-quality artistic 3D scene renderings by maintaining both global visual fidelity and local color accuracy while ensuring perspective consistency.
Our contribution is summarized as follows:

\begin{itemize}
    \item We propose a novel method to translate multi-view 2D images into stylized 3D scenes using 3D strokes based on basic primitives and spline curves whose parameters can be learned through gradient descent.
    \item We present a novel initialization scheme for the parameters in our stroke-based representation to address issues of flat gradients during optimization.
     \item  Our method supports various 3D stroke styles and can be applied for sophisticated geometry and appearance stylizations similar to those created by human painters.
\end{itemize}

%=========================================================================

\section{Related Work}
\label{sec:related_work}

%-------------------------------------------------------------------------
\vspace{-1.0mm}
\subsection{Image Painting}
\vspace{-1.0mm}

Image painting, evolving over a long history, uses vectorized strokes in various colors and styles for 2D image creation. Early methods like "Paint by Numbers"~\cite{haeberli1990paint} introduced brush parameters such as position, color, size, and direction, optimized through techniques like stroke zooming, noise adding, and color enhancement. Subsequently, more expressive forms emerged, employing slender spline curves~\cite{hertzmann1998painterly} and rectangular strokes~\cite{shiraishi2000algorithm} to better capture source image details. The development of brush libraries~\cite{seo2009painterly, zhao2011customizing} allowed for more diverse styles in paintings, notably in oil painting effects. With the advent of deep learning, techniques like generative neural networks~\cite{nakano2019neural} have further refined style representation. Advanced loss based on optimal transportation~\cite{zou2021stylized} and feed-forward paint transformer~\cite{liu2021paint} are used to improve fidelity and speed up the painting process. 
While these 2D image synthesis techniques are relatively mature, applying styles directly to 2D images from 3D scenes lacks consistency across views, which motivates our present work on using 3D strokes.

%-------------------------------------------------------------------------
\vspace{-1.0mm}
\subsection{Stylization of 3D Scenes}
\vspace{-1.0mm}

The process of 3D scene stylization involves applying the visual style of a chosen reference to a target 3D scene while preserving its inherent 3D structure. Traditional methods~\cite{sheffield1985selecting} faced challenges like limited control over perceptual factors and the inability to selectively target stylization effects. The advent of Neural Radiance Fields (NeRF) has provided a more flexible representation for 3D scene stylization. Techniques like ARF~\cite{zhang2022arf} transfer artistic features from 2D style images to 3D scenes, treating stylization as an optimization problem within the NeRF framework. Chiang et al.~\cite{chiang2022stylizing} combines NeRF's implicit representation with a hypernetwork for style transfer. StylizedNeRF~\cite{Huang22StylizedNeRF} jointly learns 2D and 3D stylization, and StyleRF~\cite{liu2023stylerf} learns high-level features in 3D space for fast zero-shot style transfer.
Our approach diverges from existing methods, which rely on additional style references and fundamentally contrast with the processes human artists use to create artworks. We focus on generating artistic images through 3D brush strokes, eliminating the need for style images. Our method more closely resembles image painting techniques than voxel-level manipulations.

%=========================================================================
\vspace{-1.0mm}
\section{Methodology}
\label{sec:method}
\vspace{-1.0mm}

In this section, we introduce our 3D scene stylization framework with stroke-based representation. Our framework mainly consists of three parts: (1) a collection of 3D strokes based on primitives and spline curves, (2) differentiable rendering and composition of strokes, (3) a stroke initialization and update scheme that stabilizes the training process.

%-------------------------------------------------------------------------
\vspace{-1.0mm}
\subsection{Overview of Stroke Field}
\label{sec:overview}
\vspace{-1.0mm}

NeRF~\cite{mildenhall2021nerf} represents the scene as the density $\sigma(\mathbf{x}) \in \mathbb{R}^{+} $ and RGB radiance $\mathbf{c}(\mathbf{x}, \mathbf{d})$, modeled by an MLP that takes the spatial coordinates $\mathbf{x} \in \mathbb{R}^3$ and view directions $\mathbf{d} \in \mathbb{R}^2$ as input.
Given a camera pose, rays $\mathbf{r}(t)=\mathbf{o} + t \mathbf{d}$ are sampled and cast from the camera's center $o \in \mathbb{R}^3$ along the direction $\mathbf{d} \in \mathbb{R}^3$ passing through each pixel of the image. The color of a ray is given by volume rendering:
\begin{equation}
    C(\mathbf{r}) = \int_{t_n}^{t_f} \sigma(\mathbf{r}(t)) \mathbf{c}(\mathbf{r}(t), \mathbf{d}) \exp(\int_{t_n}^t \sigma(\mathbf{r}(s)) ds) dt
\end{equation}

\begin{figure}[t]
  \centering
  \includegraphics[width=\linewidth]{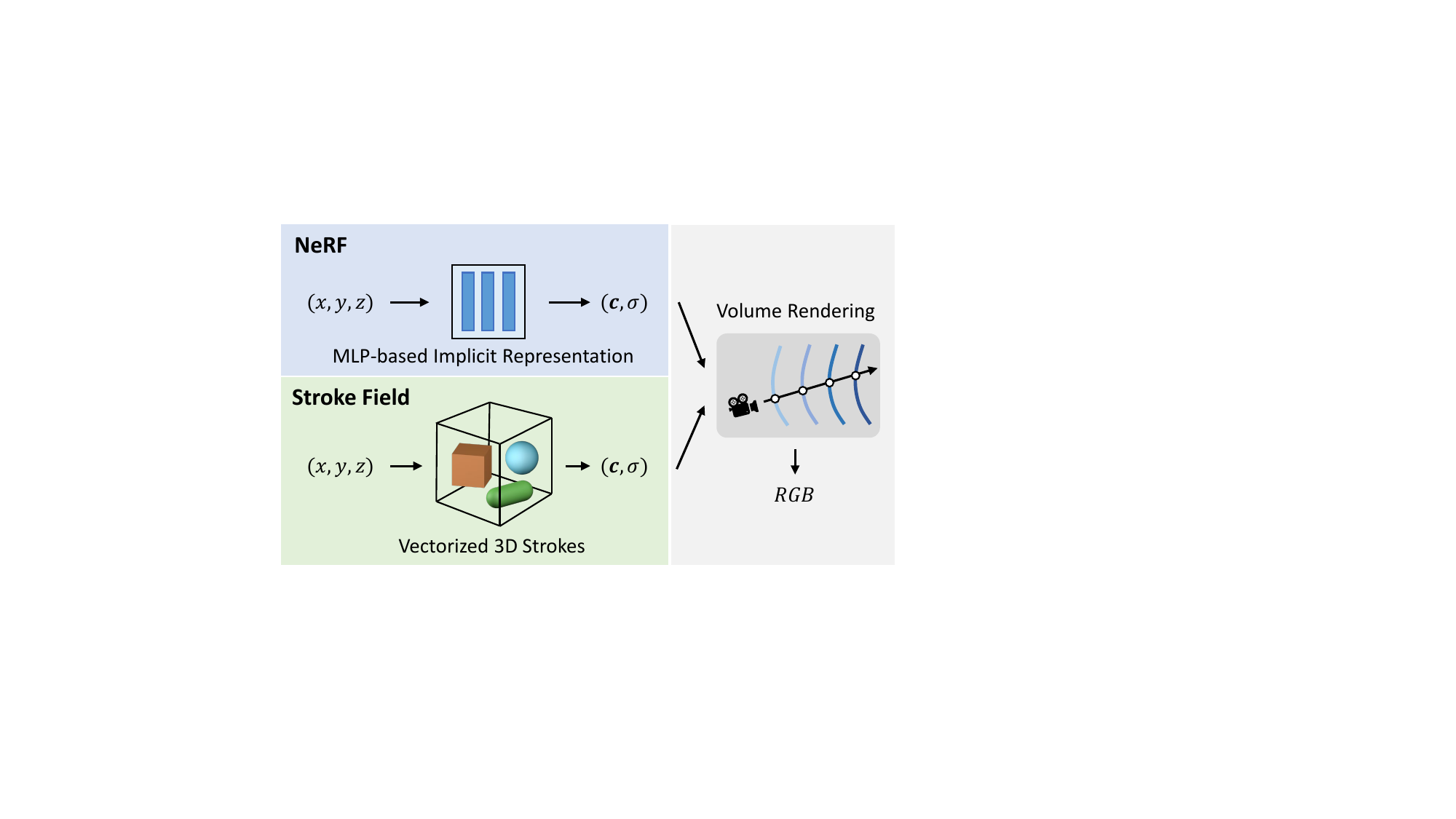}
  \caption{Our method learns a vectorized stroke field instead of MLP-based implicit representation to represent a 3D scene.}
  \label{fig:stroke_field}
  \vspace{-4.0mm}
\end{figure}

As shown in Fig.~\ref{fig:stroke_field}, while NeRF models the scene at per-voxel level with an implicit field, our method represents the scene as a set of vectorized 3D strokes. Similar to NeRF, the stroke field is defined by two spatially varying functions for density $\sigma(\mathbf{x}) \in \mathbb{R}^{+}$ and RGB color $\mathbf{c}(\mathbf{x}, \mathbf{d})$ and rendered into 2D images at a given camera pose, using the same differentiable volume rendering formula as in NeRF.

While the stroke field shares NeRF's field definition, their core formulations are fundamentally different.
Each point's density and color in the stroke field are set by 3D strokes, shaped and styled to resemble brush traces in human drawings.
This is analogous to the difference between a rasterized image and a vectorized image but in 3D space. Specifically, the geometry, appearance, and opacity of 3D strokes are given by three parameters, including shape $\mathbf{\theta_s} \in \mathbb{R}^{N \times d_s}$, color $\mathbf{\theta_c} \in \mathbb{R}^{N \times d_c}$ and density $\mathbf{\theta_\sigma} \in (\mathbb{R}^{+})^{N}$, where $N$ is the total number of strokes in the field, $d_s$ and $d_c$ are the number of parameters of the specific stroke shape and color, respectively. We combine these strokes in a differentiable way to acquire the density and color field:
\vspace{-1.0mm}
\begin{equation}
    (\sigma, \mathbf{c}) = StrokeField(\mathbf{x}, \mathbf{d};\, \mathbf{\theta_s}, \mathbf{\theta_c}, \mathbf{\theta_\sigma})
\end{equation}

%-------------------------------------------------------------------------
\vspace{-2.0mm}
\subsection{3D Strokes}
\label{sec:3d_strokes}

We first define the shape of our 3D strokes to paint them into the 3D space. The shape of a 3D stroke is essentially a volume region formed by a closed two-dimensional surface. To define such volume, we use the Signed Distance Field (SDF), a scalar value function $\mathrm{sdf}(\mathbf{x}) \to s \in \mathbb{R}$ that gives the signed distance from a point to the closed surface to describe the stroke shape in 3D. The sub-space $\{\mathbf{p} \in \mathbb{R}^3 \; | \; \mathrm{sdf}(\mathbf{p}) \leq 0\}$ enclosed by the zero level-set is the stroke volume. We construct two types of strokes based on basic geometric primitives and spline curves respectively.

\vspace{-2.0mm}
\subsubsection{Basic Primitives}
\label{sec:basic_primitives}
\vspace{-2.0mm}

Our first category of 3D strokes is based on common geometric primitives such as spheres, cubes, etc. For simplicity, we normalize these geometries at the origin of the unit space, apply a transformation matrix to the unit geometries, and acquire their SDFs in the scene space. The unit geometry can optionally contain a shape parameter $\theta_s^\text{basic} \in \mathbb{R}^{d_\text{basic}}$ describing its transformation-agnostic shape feature, where $d_\text{basic}$ is the number of basic shape parameters. The SDF of unit primitive is defined as $\mathrm{sdf}_\text{unit}: (\mathbf{\hat{p}}, \theta_s^\text{basic}) \to s \in \mathbb{R}$, where $\mathbf{\hat{p}}$ is a 3D point in the unit space. We list the common unit geometric primitives in Tab.~\ref{tab:basic_sdfs}.

\begin{table}[t]
    \centering
    \resizebox{1.00\columnwidth}{!}{
    \begin{tabular}{c|c|p{5.6cm}}
    \toprule
    Primitive & Params & SDF formula \\
    \midrule
    Sphere & None & $\Vert \mathbf{p} \Vert_2 - 1$ \\
    \midrule
    Cube & None & $\min(\max(\mathbf{q}_x,\mathbf{q}_y,\mathbf{q}_z),0) + \Vert \max(\mathbf{q},\mathbf{0}) \Vert_2$, where $\mathbf{q} = |\mathbf{p}| - \mathbf{1}$ \\
    \midrule
    Tetrahedron & None & $(\max(|\mathbf{p}_x+\mathbf{p}_y|-\mathbf{p}_z,|\mathbf{p}_x+\mathbf{p}_y|+\mathbf{p}_z) - 1) / \sqrt{3}$ \\
    \midrule
    Octahedron & None & $(\Vert \mathbf{p} \Vert_1 - 1) / \sqrt{3}$ \\
    \midrule
    Round Cube & $r$ & $\min(\max(\mathbf{p}_x,\mathbf{p}_y,\mathbf{p}_z),0) + \Vert \max(\mathbf{p},\mathbf{0}) \Vert_2 - r $ \\
    \midrule
    Triprism & $h$ & $\max(|\mathbf{p}_y| - h, \max(|\mathbf{p}_x| * \sqrt{3} / 2 + \mathbf{p}_z / 2, -\mathbf{p}_z) - 0.5)$ \\
    \midrule
    Capsule Line & $h,r_\delta$ & $\Vert \mathbf{p} - [0,\min(\max(\mathbf{p}_y,-h),h),0] \Vert_2 - r_\delta \min(\max((0.5*(\mathbf{p}_y+h)/h,0),1) - 1$ \\
    \bottomrule
    \end{tabular}
    }
    \caption{SDF formula for common unit geometry primitives.}
    \label{tab:basic_sdfs}
    \vspace{-4.0mm}
\end{table}

We apply a transformation matrix $\mathbf{T} \in \mathbb{R}^{4 \times 4}$ to transform primitives from the unit coordinates $\mathbf{\hat{p}}$ to the shared scene coordinates $\mathbf{p} = \mathbf{T} \mathbf{\hat{p}}$. The parameters of transformation are composed of a translation $\mathbf{t} \in \mathbb{R}^3$, a rotation described by Euler angle $\mathbf{r} \in \mathbb{R}^3$, and a uniform or anisotropic scale $\mathbf{s} \in \mathbb{R}^3$.
To acquire the SDF of primitives in the scene space, we inversely transform the position to the unit space and query the unit SDF:
\vspace{-2.0mm}
\begin{equation}
    \mathrm{sdf}(\mathbf{p}; \theta_s) = \mathrm{sdf}_\text{unit}(\mathbf{T}(\mathbf{t}, \mathbf{r}, \mathbf{s})^{-1}\mathbf{p},\theta_s^\text{basic}),
\vspace{-2.0mm}
\end{equation}
where $\theta_s=\{\theta_s^\text{basic}, \mathbf{t}, \mathbf{r}, \mathbf{s}\}$ is the final shape parameters for primitive-based 3D strokes. In practice, we may use a subset of translation, rotation, and scale to combine different styles of primitive strokes. We list all the primitive-based 3D strokes in the supplementary. More primitive can be easily added given a defined unit SDF.

\vspace{-2.0mm}
\subsubsection{Spline Curves}
\label{sec:spline_curves}
\vspace{-2.0mm}

In addition to basic primitives, we use volumetric 3D curves with a given radius to simulate the trace of strokes in human paintings. The curves are defined by parametric 3D splines in the scene space: $C: (t, \theta_s^\text{curve}) \to \mathbf{x} \in \mathbb{R}^3$, where $t \in [0, 1]$ is the interpolation parameter, and $\theta_s^\text{curve} \in \mathbb{R}^{d_c}$ is the parameter of the 3D spline. To simulate the brushstroke effects, we define two different radii $r_a, r_b \in \mathbb{R}^+$ at the two endpoints of the spline curve respectively, and interpolate the radius at any position on the curve smoothly based on $t$ as $r(t; r_a, r_b) = r_a (1-t) + r_b t$.

We utilize common polynomial splines, including quadratic and cubic Bézier and Catmull-Rom splines, to define 3D curves. To compute the SDF of a 3D curve, we need to solve for the value of $t$ corresponding to the nearest point on the spline curve for any point $\mathbf{p} \in \mathbb{R}^3$ in space. While there exists an analytic solution for quadratic splines, it is difficult to directly solve for the nearest point's $t$ value for cubic or more complex splines in a differentiable way. 

Therefore, we use a general approximation method to compute the SDF: we uniformly sample $K+1$ positions on the curve to get $K$ line segments, and calculate the distance from these segments to the query point respectively, thereby obtaining an approximate nearest point. This algorithm can be applied to any parametric curve spline and allows simple control over the trade-off between computational complexity and accuracy.
With the found $t$ value of the nearest point on the 3D spline to query position $\mathbf{p}$ as $t^*$, the SDF of the 3D curve is defined as
\vspace{-1.0mm}
\begin{equation}
    \mathrm{sdf}(\mathbf{p}; \theta_s) = \Vert \mathbf{p} - C(t^*,\theta_s^\text{curve}) \Vert_2 - r(t^*; r_a, r_b)
\vspace{-1.0mm}
\end{equation}
where $\theta_s=\{\theta_s^\text{curve}, r_a, r_b\}$ is the final shape parameters for spline-based 3D strokes. We leave the details of the nearest point finding algorithm in the supplementary.

%-------------------------------------------------------------------------
\subsection{Differentiable Rendering of 3D Strokes}
\label{sec:diff_rendering}

With the SDF of 3D strokes defined in Sec.~\ref{sec:3d_strokes}, we now convert it into the density field $\sigma(\mathbf{x})$ and color field $\mathbf{c}(\mathbf{x}, \mathbf{d})$ for differentiable rendering.

\begin{figure}[t]
  \centering
  \includegraphics[width=\linewidth]{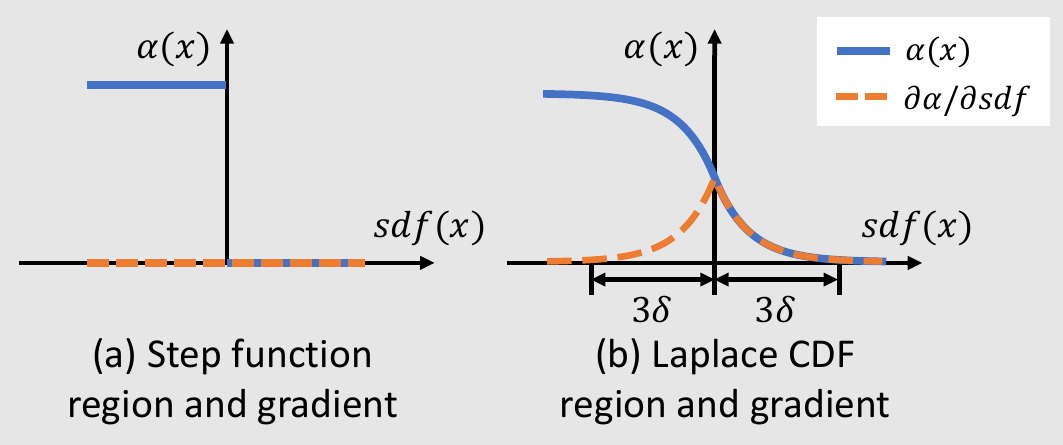}
  \caption{Differential region function by approximating step function with the CDF of Laplace distribution. The parameter $\delta$ controls the width of the transitional area with respect to the SDF of 3D strokes, where a larger $\delta$ leads to smoother gradient at the cost of making the shape boundary blurrier.}
  \label{fig:region_function}
  \vspace{-4.0mm}
\end{figure}

\vspace{-2.0mm}
\paragraph{Deriving Density Field from SDF.}
Theoretically, we consider the inner region of a 3D stroke with an SDF value less than or equal to zero. Therefore, we can define an indicator region function $\alpha(\mathbf{x}) \in [0,1]$ based on whether a point is inside the SDF:
\vspace{-2.0mm}
\begin{equation}
    \alpha(\mathbf{x}) = \left \{ \begin{aligned} & 1,\; \mathrm{sdf}(\mathbf{x}) \leq 0 \\ & 0,\; otherwise \end{aligned} \right.
\vspace{-2.0mm}
\end{equation}
Assuming uniform density inside one stroke, we define the density field of each 3D stroke as
\vspace{-2.0mm}
\begin{equation}
    \sigma(\mathbf{x}) = \theta_\sigma \alpha(\mathbf{x})
\vspace{-2.0mm}
\end{equation}
where $\theta_\sigma \in \mathbb{R}^+$ is the density parameter of a 3D stroke. However, due to the discontinuity of the indicator function, the gradient of the density field w.r.t. the shape parameters of the SDF is zero, thus hindering the use of losses to optimize the 3D stroke. To render the 3D stroke in a differentiable way, we aim to derive a smooth approximation of the region function to gain non-zero gradients regarding the shape parameters. %Since the discrete version of the region function can be considered as a step function of the SDF value, 
We use the Laplace cumulative distribution function (CDF) to approximate the discrete region function, similar to VolSDF~\cite{yariv2021volume}:
\vspace{-1.0mm}
\begin{equation}
    \alpha(\mathbf{x}) = \left \{ \begin{aligned} & 1 - \frac{1}{2}\exp(\mathrm{sdf}(\mathbf{x}) / \delta),\; \mathrm{sdf}(\mathbf{x}) \leq 0 \\ & \frac{1}{2} \exp(-\mathrm{sdf}(\mathbf{x}) / \delta),\; otherwise \end{aligned} \right.
\vspace{-1.0mm}
\end{equation}
where $\delta$ controls the width of the smooth transitional interval, as shown in Fig.~\ref{fig:region_function}. This definition ensures that all points in space have non-zero derivatives w.r.t. the shape parameters, although the gradient would be too small to achieve meaningful optimization excluding the near region besides the zero level-set boundary of the SDF. This approximation approaches the discrete indicator function when $\delta \to 0$, while larger $\delta$ allows smooth gradients flow in a larger area near the boundary, at the cost of making the shape boundary more blurred. 

% MOVE TO supplementary material ?
% We demonstrate the gradient of different choices of $\delta$ in Fig.~\ref{} \macc{to be added}.

\vspace{-2.0mm}
\paragraph{Adaptive Choice of $\delta$ based on Cone Tracing.}
Since the pixel corresponding to each ray has a specific planner size during rendering, the pinhole camera model casts a frustum with non-zero volume instead of a ray. The frustum far away from the camera will cover a large region of the scene, while the near frustum will only affect a small region. Using a uniform $\delta$ across the entire scene to render the 3D strokes will make the near objects over-blurry, and far objects lack enough gradients to optimize the stroke shape. To solve this issue, we adopt a similar cone tracing strategy as in MipNeRF~\cite{barron2021mip}, and adjust the $\delta$ value of the region function adaptively based on the cone size. For simplicity, we use isotropic spheres instead of anisotropic multivariate Gaussians to approximate the volume of the cone.
Specifically, assuming the radius of each pixel is $\dot{r}$, we calculate the radius of a sphere at sample points as $r = \dot{r} t f^{-1}$, where $t$ is the un-normalized ray distance, and $f$ is the focal length. We then compute the adaptive $\delta$ as $\delta = (k_\delta r)^{-1}$, where $k_\delta$ is a hyper-parameter controlling how much to ``dilate" the boundary.
Moreover, as the scale of $\delta$ is defined in scene space, for basic primitives with a transformation, we further adjust $\delta$ based on the scaling factor of the inverse transform.

\vspace{-2.0mm}
\paragraph{Color Field.}
We mainly use constant color for each stroke, which is irrelevant to the position and view direction and can be expressed as $\mathbf{c}(\mathbf{x}, \mathbf{d}) = \theta_\mathbf{c}$, where $\theta_\mathbf{c} \in \mathbb{R}^3$ is the RGB color parameter. More diverse visual effects can be achieved by spatially varying color fields and joint modification of the density and color. 
% We demonstrate the application of using textured color based on shape-conditioned texture coordinates in Sec.~\ref{sec:app_texture}. 
The color field can be readily expanded to support view-dependent effects by replacing the RGB color with spherical harmonics, however, we have found that the view dependency does not contribute much to the visual effect of stroke-based scenes in practice.

%-------------------------------------------------------------------------
\subsection{Composition of 3D Strokes}
\label{sec:stroke_composition}

Given the density and color field of individual 3D strokes, we now compose them into a stroke field for rendering the full 3D scene. In the procedure of image painting, different strokes are painted onto the canvas sequentially, and new strokes are overlayed on the old strokes. This behavior is simulated using the alpha composition. Similarly, we use an ``overlay" method to compose multiple 3D strokes.

Suppose we have the density fields $\{\sigma_i(\mathbf{x})\}$ and color fields $\{\mathbf{c}_i(\mathbf{x})\}$ of $N$ strokes, where $i \in \{1,2,\cdots,N\}$. We use the region function defined in Sec.~\ref{sec:diff_rendering} as the blending weight for overlay. Given $\{\alpha_i(\mathbf{x})\}$ as the region function of each 3D stroke, the density and color of the final stroke field are expressed as
\vspace{-2.0mm}
\begin{equation}
\begin{split}
    \sigma(\mathbf{x}) &= \sum_i^n \sigma_i(\mathbf{x}) \alpha_i(\mathbf{x}) T_i(\mathbf{x}), \\
    \mathbf{c}(\mathbf{x}) &= \frac{\sum_i^n \mathbf{c}_i(\mathbf{x}) \alpha_i(\mathbf{x}) T_i(\mathbf{x})}{1 - T_n(\mathbf{x})},
\end{split}
\end{equation}
\vspace{-2.0mm}
where
\vspace{-2.0mm}
\begin{equation}
    T_i(\mathbf{x})=\prod_j^i (1 - \alpha_j(\mathbf{x})).
\vspace{-2.0mm}
\end{equation}
Note that we normalize $\mathbf{c}(\mathbf{x})$ using the total weight of all region functions, making the final color unaffected by the number of strokes.

The ``overlay" composition is sensitive to the painting order of multiple strokes, i.e., the 3D strokes painted later will have larger blending weights than the previously painted strokes. Normally this is the behavior we desire, however, we sometimes would like the final outcome irrelevant to the painting order. The simplest way is choosing the 3D stroke with the maximum $\alpha(\mathbf{x})$ as: $\sigma(\mathbf{x})=\sigma_i(\mathbf{x}), \; \mathbf{c}(\mathbf{x})=\mathbf{c}_i(\mathbf{x})$, where $i=\arg \max_j\{\alpha_j(\mathbf{x})\}$. This ``max" composition will only attribute the gradient of the density field to the nearest 3D stroke. A smoother way is to use Softmax for computing weights as: $\sigma(\mathbf{x})=\sigma_i(\mathbf{x}) \omega_i(\mathbf{x}), \; \mathbf{c}(\mathbf{x})=\mathbf{c}_i(\mathbf{x}) \omega_i(\mathbf{x})$, where $\omega_i(\mathbf{x})=\frac{\exp(\alpha_i(\mathbf{x}) / \tau)}{\sum_j^n \exp(\alpha_j(\mathbf{x}) / \tau)}$, and $\tau$ is a hyper-parameter that controls the blending smoothness. We analyze the effect of different compositions in Sec.~\ref{sec:abl_composition}.

%-------------------------------------------------------------------------
\begin{figure*}[t]
  \centering
  \includegraphics[width=\linewidth]{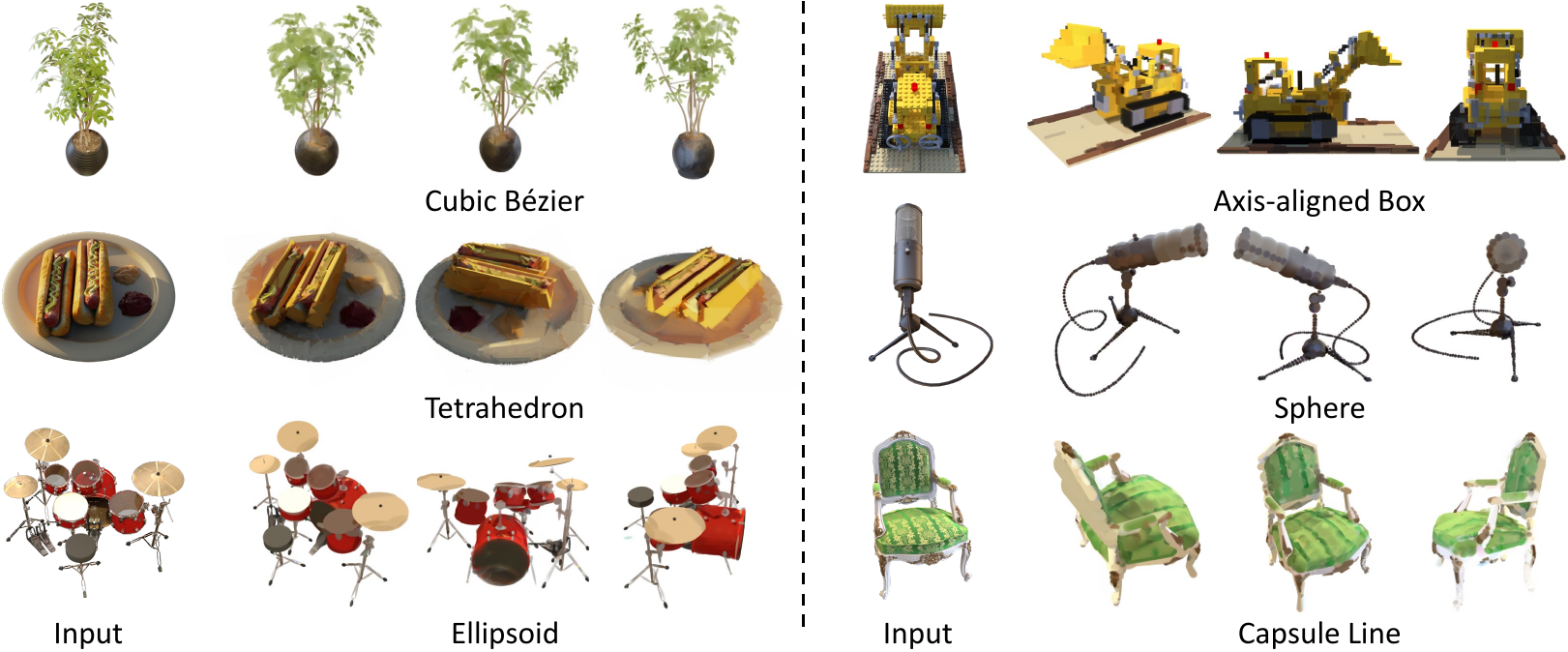}
  \caption{Novel view synthesis results of various 3D strokes on the synthetic scenes. Each scene contains 500 strokes. Our vectorized stroke representation is able to recover the 3D scenes with high fidelity while maintaining a strong geometric style defined by the strokes.}
  \label{fig:results_blender}
  \vspace{-5.0mm}
\end{figure*}

\subsection{Training Strategy}
\label{sec:train_strategy}

Now we have completed our rendering framework for 3D strokes. We can render 2D images with a given camera pose from parameters of $N$ strokes $\{\theta_s,\theta_c,\theta_\sigma\}_i,\; i \in \{1,2,\cdots,N\}$, and use gradient descent to optimize these 3D stroke parameters. However, in the previous analysis, we mentioned that the magnitude of gradients vanishes for faraway points from the existing boundary of strokes, thus leaving shape parameters unable to gain enough optimization. To this end, we propose a training scheme for better stroke initialization and update.

\vspace{-2.0mm}
\paragraph{Error Field.}
When adding 3D strokes to an existing scene, we wish to initialize the new stroke at the location where it is most needed. Therefore, during the training of the stroke field, we additionally train an error field $e(\mathbf{x}) \in \mathbb{R}^+$ to describe the ``error rate" at each position of the reconstructed scene, with the region of highest error considered to be the locations most in need of additional 3D strokes. The error value $E(\mathbf{r})$ along each ray $\mathbf{r}$ is obtained by the modified volume rendering formula:
\begin{equation}
    E(\mathbf{r})=\int_{t_n}^{t_f} e(\mathbf{r}(t)) \exp(\int_{t_n}^t e(\mathbf{r}(s)) ds) dt.
\end{equation}
We define the error as the squared difference in color between the stroke field rendering result and the ground truth. Moreover, we prefer the error field to conservatively estimate errors, therefore we use a coefficient $k > 1$ to amplify the loss where the error is underestimated. Assuming the rendering color of stroke field is $C(\mathbf{r})$, and the GT color is $C_\text{gt}(\mathbf{r})$, then the training loss for the error field is
\begin{equation}
    \mathcal{L}_{err}=|d| k^{\max(-sgn(d),0)} ,\; d=E(\mathbf{r}) - \Vert C(\mathbf{r})-C_\text{gt}(\mathbf{r}) \Vert_2,
\end{equation}
where $sgn(\cdot)$ is the sign function. At the same time, for views that are less trained, we apply a regularization loss to the error values. For $N$ total sampling points $\{\mathbf{x}_i\}$ in a batch, the regularization loss is $\mathcal{L}_{err\,reg}=\sum_i^N e(\mathbf{x}_i)$.

\vspace{-2.0mm}
\paragraph{Stroke Initialization and Update.}
When training a scene with $N$ strokes, we randomly initialize $N_\text{start}$ strokes and then incrementally add the remaining strokes step by step to the scene. We also decay the initial size of strokes based on the current number of strokes, i.e., we progressively use smaller strokes as the number of existing strokes increases. When initializing the $i$-th stroke, we uniformly sample $M$ coordinates $\{\mathbf{p}_i\}$ within the bounding box of the scene, calculate the error values $\{e_i\}$ in the error field, and select the sampling point with the highest error as the initial position for the 3D stroke. For primitives, we set their translation vector to this initial position. For spline curves, we set their control points to this initial position plus a random offset vector sampled within the current stroke size volume. This initialization scheme, guided by the error field, effectively mitigates optimization issues caused by flat gradients. Additionally, we reinitialize strokes with near-zero density by resampling the current error field. This simple technique recycles some of the brush strokes that might have been trapped in local minima due to poor initialization.

\vspace{-2.0mm}
\paragraph{Training Objective.}
Similar to NeRF, we use multi-view images as the primary supervision. We randomly sample rays from images in the training set and optimize the error between the rendering result of each ray and the ground truth. We use the Charbonnier distance for color loss:
\begin{equation}
    \mathcal{L}_{color}= \sqrt{\Vert C(\mathbf{r}) - C_{gt}(\mathbf{r}) \Vert_2^2+\epsilon}
\end{equation}
For scenes with a ground truth mask, we also impose mask supervision based on the ray's opacity $O(\mathbf{r}) = \exp(\int_{t_n}^{t_f} \sigma(\mathbf{r}(t)) dt)$ and the ground truth $M(\mathbf{r}) \in \{0,1\}$,
\vspace{-1.0mm}
\begin{equation}
    \mathcal{L}_{mask} = \sqrt{\Vert O(\mathbf{r}) - M_{gt}(\mathbf{r}) \Vert_2^2+\epsilon}
\vspace{-1.0mm}
\end{equation}
Additionally, we regularize the density parameter of $N$ strokes as $\mathcal{L}_{den\,reg}= \sum_i^N |\theta_\sigma^i|$. Combining the above losses, our total loss is given by:
\vspace{-2.0mm}
\begin{equation}
\begin{split}
    \mathcal{L}&=\lambda_{color}\mathcal{L}_{color}+\lambda_{mask}\mathcal{L}_{mask} \\
    &+\lambda_{den\,reg}\mathcal{L}_{den\,reg}+\lambda_{err}\mathcal{L}_{err}+\lambda_{err\,reg}\mathcal{L}_{err\,reg}
\end{split}
\vspace{-2.0mm}
\end{equation}

%=========================================================================

\begin{figure*}[!t]
  \centering
  \includegraphics[width=1.0\linewidth]{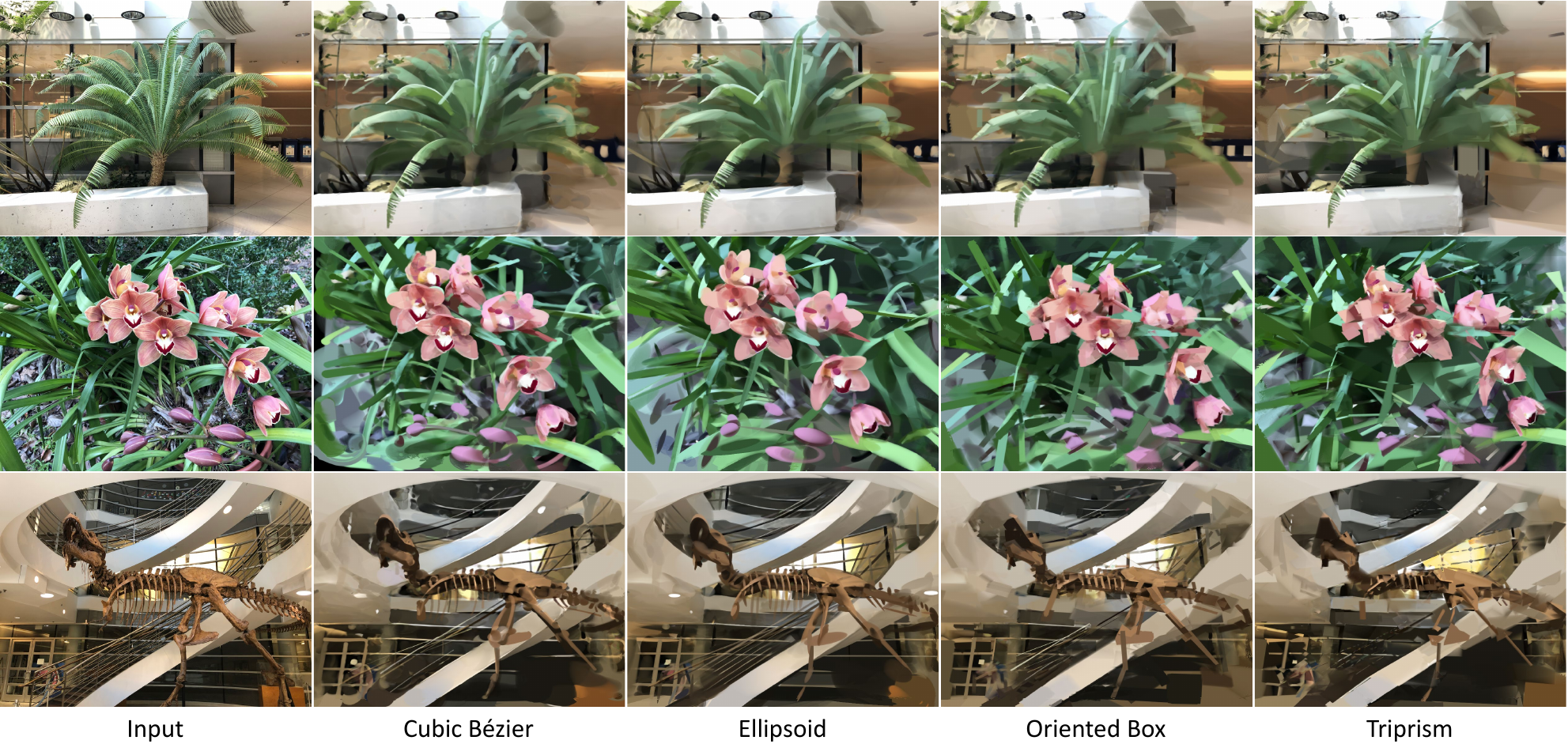}
  \caption{Results of different types of 3D strokes on the face-forwarding scenes. Each scene contains 1000 strokes.}
  \label{fig:results_llff}
  \vspace{-2.0mm}
\end{figure*}

\begin{figure*}[!t]
  \centering
  \includegraphics[width=1.0\linewidth]{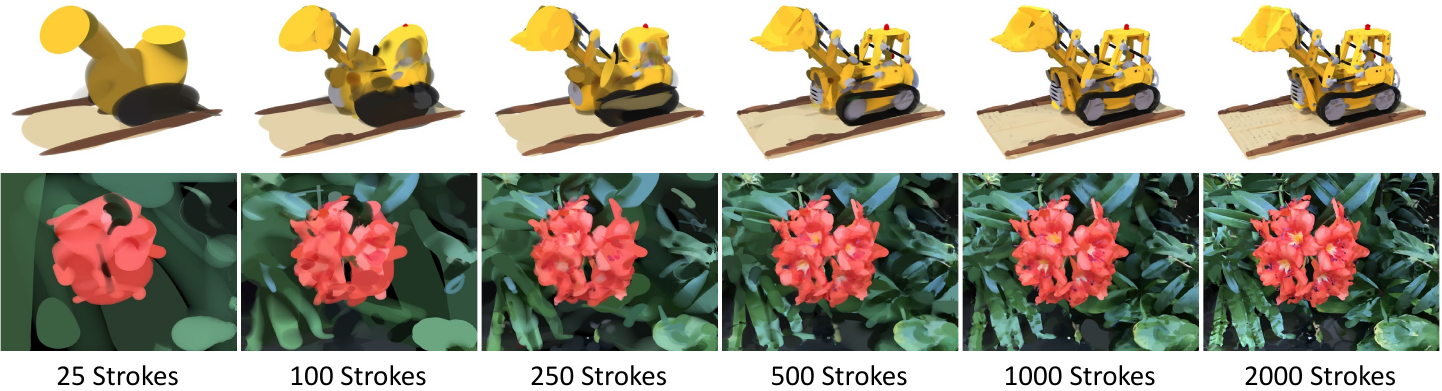}
  \caption{Results with different stroke numbers. By changing the total number of strokes, our stroke-based representation can capture both the abstraction and fine-grained details of the original scenes.}
  \label{fig:progress}
  \vspace{-3.0mm}
\end{figure*}

%=========================================================================

\section{Experiments}
\label{sec:experiments}

We test our method on both synthetic objects from Blender~\cite{mildenhall2021nerf} and face-forwarding scenes from LLFF~\cite{mildenhall2019local}. Each dataset comprises dozens to hundreds of multi-view images with given camera poses. We show our reconstruction results based on various types of vectorized 3D strokes, compare the stylization results of our method with other image-to-painting methods, and conduct an ablation study.

%-------------------------------------------------------------------------

\subsection{Stroke-based scene reconstruction}
\label{sec:exp_recon}

Fig.~\ref{fig:results_blender} shows a sample of the stylized 3D synthetic scenes reconstructed by our method with several types of 3D strokes. Our method is able to synthesize 3D scenes based on vectorized 3D strokes, providing strong geometry and appearance stylization while maintaining the abstracted shapes and colors of the original scenes. Fig.~\ref{fig:results_llff} compares reconstruction results of different types of 3D strokes on the same scene. Cubic Bézier and Ellipsoid generally have the best reconstruction fidelity compared to other types of strokes. We list the quantitative metrics of different strokes in the supplementary material.
% supplementary material
% We also list the quantitative metrics of our reconstructed scenes in Tab.\ref{}.
Fig.~\ref{fig:progress} demonstrates the abstract-to-detail painting results by varying the number of total 3D strokes that are used to reconstruct the scene. We can observe that the lower number of strokes approximates the original scene with a strong shape-related geometric style, while the higher number of strokes leads to reconstruction results with higher fidelity.

\begin{figure}[!t]
  \centering
  \includegraphics[width=\linewidth]{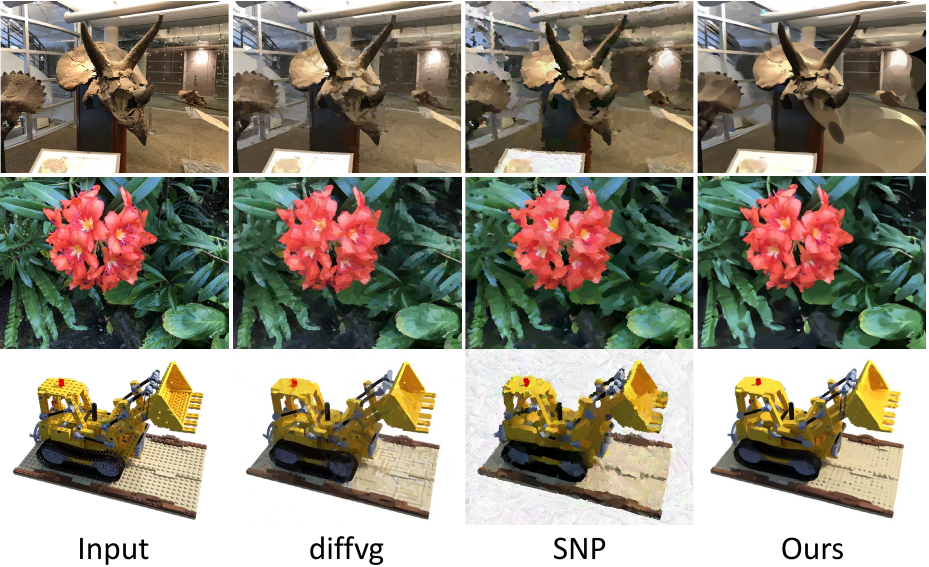}
  \caption{Comparison with 2D image-to-painting methods.}
  \label{fig:comparison}
  \vspace{-5.0mm}
\end{figure}

%-------------------------------------------------------------------------
\subsection{Comparison with other methods}
\label{sec:exp_comparison}

Our method lies in a different category of stylization compared to the common NeRF-based stylization works, which usually take a reference image as input, and transfer the textural style onto the appearance of a trained radiance field. Our method does not require extra reference images, instead, the style of our method is intrinsic and contained in the specification of various 3D strokes.

We compare our method with 2D image-to-painting methods. Fig.~\ref{fig:comparison} shows the vectorized reconstruction results of our method and other 2D vectorized image representation methods, including diffvg~\cite{li2020differentiable} and Stylized Neural Painting~\cite{zou2021stylized}.
We use multi-view images to train our 3D stroke representation, while for 2D methods we select images of the specific viewpoints and apply the vectorization tracing. It can be observed that all these vectorized methods are able to reconstruct original images with obvious stroke styles. Nevertheless, our method can recover more details and strictly maintain the multi-view consistency across different views, which is guaranteed by its 3D representation. The results are best viewed in our supplementary video.

%-------------------------------------------------------------------------
\subsection{Ablation studies}
\label{sec:exp_ablation}

\begin{figure}[!t]
  \centering
  \includegraphics[width=\linewidth]{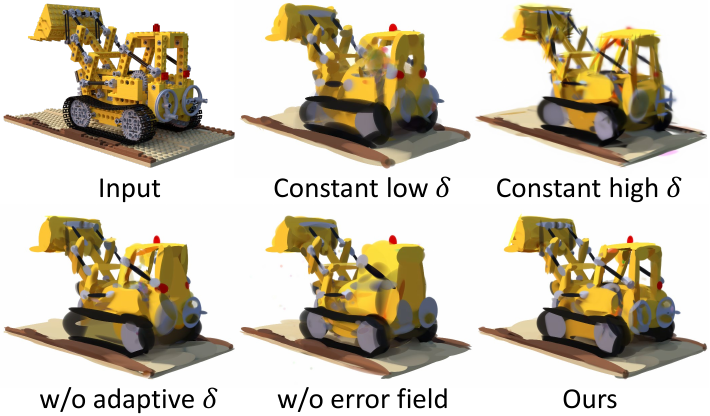}
  \caption{Comparison of ablation study results.}
  \label{fig:ablation}
  \vspace{-2.0mm}
\end{figure}

\vspace{-1.0mm}
\paragraph{Use of adaptive $\delta$ in region function.} \label{sec:abl_delta}
We investigate the effects of employing an adaptive $\delta$ in differentiable rendering. The results in Tab.~\ref{tab:ablation} show that omitting adaptive $\delta$ leads to reduced performance in novel view synthesis. Additionally, Fig.~\ref{fig:ablation} demonstrates that constant $\delta$ values, whether low or high, result in sub-optimal or overly blurry scenes.
%We explore the impact of using an adaptive $\delta$ in differential rendering. The results in Tab.~\ref{tab:ablation} indicate the exclusion of adaptive $\delta$ results in lower novel view synthesis metrics. Meanwhile, as shown in Fig.~\ref{fig:ablation}, using either a low or high constant $\delta$ leads to either insufficiently optimized or excessively blurry scenes.

\vspace{-4.0mm}
\paragraph{Use of error field.} \label{sec:abl_errorfield}
In Sec.~\ref{sec:diff_rendering}, we note that loss gradients are concentrated near the 3D stroke boundary, creating many local minima that heavily influence optimization based on shape initialization. Tab.~\ref{tab:ablation} demonstrates that using the error field for stroke shape initialization significantly improves scene reconstruction fidelity.
%As mentioned in Sec.~\ref{sec:diff_rendering}, the gradient of loss w.r.t. the shape parameters are concentrated near the boundary of the 3D stroke surface, thus the loss landscape contains a lot of local minima which can cause the optimization process to be largely influenced by the shape initialization. We compare the effectiveness of using the error field for better stroke initialization in Tab.~\ref{tab:ablation}. The results indicate that using the error field results in better fidelity in the reconstructed scenes.

\begin{figure}[!t]
  \centering
  \includegraphics[width=\linewidth]{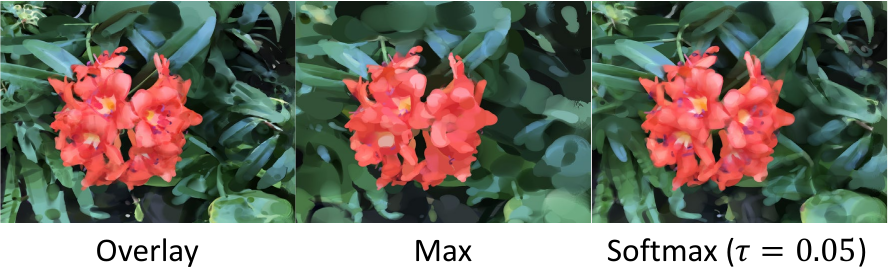}
  \caption{Comparison of stroke composition methods.}
  \label{fig:ablation_comp}
  \vspace{-5.0mm}
\end{figure}

\vspace{-4.0mm}
\paragraph{Choice of composition function.} \label{sec:abl_composition}
We compare different composition approaches in Sec.~\ref{sec:stroke_composition}. We can observe in Fig.~\ref{fig:ablation_comp} that the `overlay' composition leads to the best reconstruction quality, while the `softmax' composition also reconstructs the scene well. The latter can be chosen if order invariant painting is required.

\begin{table}
  \centering
  \resizebox{0.9\columnwidth}{!}{%
  \begin{tabular}{lccc}
    \toprule
    Method & PSNR$\uparrow$ & SSIM$\uparrow$ & LPIPS$\downarrow$ \\
    \midrule
    Ours (constant low $\delta$) & $21.30$ & $0.793$ & $0.177$ \\
    Ours (constant high $\delta$) & $21.86$ & $0.796$ & $0.235$ \\
    Ours (w/o adaptive $\delta$) & $22.50$ & $0.815$ & $0.179$ \\
    Ours (w/o error field) & $21.79$ & $0.811$ & $0.192$ \\
    Ours (full) & $\mathbf{23.13}$ & $\mathbf{0.825}$ & $\mathbf{0.171}$ \\
    \bottomrule
  \end{tabular}}
  \caption{Novel view synthesis metrics of ablation studies.}
  \label{tab:ablation}
  \vspace{-2.0mm}
\end{table}

%-------------------------------------------------------------------------
\subsection{Applications}
\label{sec:exp_app}
\vspace{-1.0mm}

We explore various applications based on our vectorized 3D scene representations, including color transfer and text-driven zero-shot scene drawing.

\begin{figure}[!t]
  \centering
  \includegraphics[width=\linewidth]{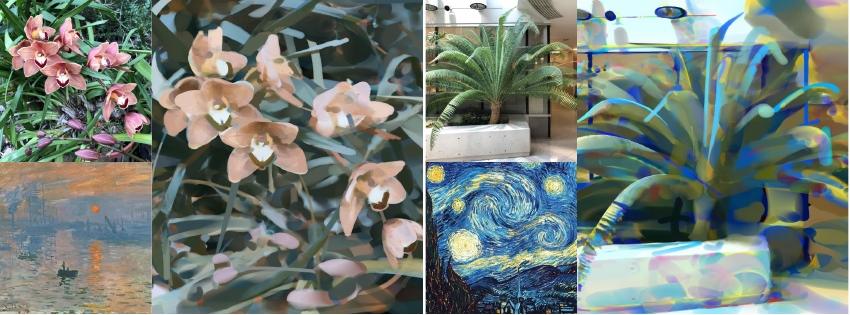}
  \caption{Color transfer results. The left column is the input and style target images respectively.}
  \label{fig:colortrans}
  \vspace{-2.0mm}
\end{figure}

\vspace{-3.0mm}
\paragraph{Color Transfer.} \label{sec:app_color}
As the 3D stroke representation has separate shape and appearance parameters, we can fix the geometry and only fine-tune the color parameter to achieve color transfer effects. We adopt the style loss in common style transfer works, which matches the gram matrix of the feature map outputs from the 4-th and 9-th layers of a pre-trained VGG network. The reference style images and color transfer results are demonstrated in Fig.~\ref{fig:colortrans}.

% \dhb{
% \vspace{-2.0mm}
% \paragraph{Textured Strokes.} \label{sec:app_texture}
% Although we mainly use a constant color function in drawing the 3D strokes, our method can be easily extended to drawn strokes with textural details by modifying the color field of each stroke. To add textural details in the stroke-based scenes, we build a local coordinate system in the defined 3D stroke according to their shape and use it to compute UV positions for mapping colors from custom textures.
% }

\vspace{-3.0mm}
\paragraph{Text-driven Scene Drawing.} \label{sec:app_generation}
We also explore using the vectorized 3D stroke representation to achieve scene creation tasks under a text-guided zero-shot generation framework. We use CLIP~\cite{radford2021learning}, a vision-language model that embeds the 2D images and text prompts into the same embedding space. Following the generative setup in DreamField~\cite{jain2021dreamfields}, we sample camera poses following a circular path around the scene origin and render a large image patch, and then we optimize the distance between CLIP embeddings of the synthesized image and the text prompt. The generated 3D drawings of different objects are shown in Fig.~\ref{fig:text_drawing}. We leave more details of the training setup in the supplementary material.

\begin{figure}[!t]
  \centering
  \includegraphics[width=\linewidth]{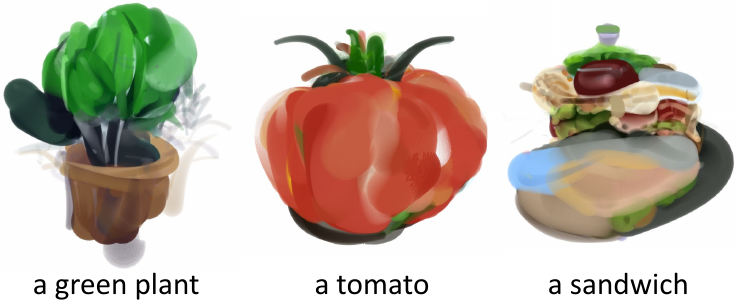}
  \caption{Text-driven 3D drawings of common objects with 300 strokes of cubic Bézier curve.}
  \label{fig:text_drawing}
  \vspace{-4.0mm}
\end{figure}

%=========================================================================
\vspace{-1.0mm}
\section{Discussion}
\label{sec:discussion}
\vspace{-1.0mm}

We present a novel method to stylize a 3D scene from multi-view 2D images. Different from NeRF-based representations, our method represents the scene as vectorized 3D strokes, mimicking human painting during scene reconstruction process. We demonstrate that this stroke-based representation can successfully stylize 3D scenes with large geometry and appearance transformations, which was not achieved with previous NeRF stylization approaches. 

\vspace{-3.0mm}
\paragraph{Limitations and future works.} 
Our method uses a stroke setting that demands manual effort to design stroke shapes and appearances. The 3D strokes can be further learned with a generative framework to create a variety of stroke types, like ink and oil brushes, with generated and more detailed SDFs.
Moreover, our method may require numerous strokes to represent very complex scenes, partly due to the existence of many local minima during the optimization. Incorporating a globally aware loss, like the optimal transport loss in~\cite{zou2021stylized} into 3D space, may enhance the convergence efficiency of our method, which we leave as future work.

\vspace{-3.0mm}
\paragraph{Acknowledgment.}
This work was supported by the National Natural Science Foundation of China (Project Number: 62372025 and 61932003) and the Fundamental Research Funds for the Central Universities. 

%=========================================================================

{
    \small
    \bibliographystyle{ieeenat_fullname}
    \bibliography{main}
}
%\end{bibunit}

\clearpage
\setcounter{page}{1}
\maketitlesupplementary

This supplementary document provides additional details of the 3D strokes in Sec.~\ref{sec:stroke_details}, implementations in Sec.\ref{sec:impl_details}, 
%visualization of the error field in Sec.~\ref{sec:vis_errorfield}, 
additional comparisons of various 3D strokes in Sec.~\ref{sec:stroke_metrics}, and application training setups in Sec.~\ref{sec:app_setups}. Please also watch our accompanying video for an animated visualization of stylization results.

\section{Details of 3D Strokes}
\label{sec:stroke_details}

\subsection{Transformation of Basic Primitives}

In Sec.~\ref{sec:basic_primitives} of the main paper, we use a transformation matrix to map the coordinates in the shared scene space into the canonical space of each unit signed distance field. Here we provide the construction details of the transformation matrix. Given a translation vector $\mathbf{t} = (t_x, t_y, t_z)$, an Euler angle rotation vector $\mathbf{r} = (r_x, r_y, r_z)$, and a scale vector $\mathbf{s} = (s_x, s_y, s_z)$, we first construct the matrices for each term respectively, then combine them in the order of scale, rotation, and translation to get the final transformation matrix $M$:
\begin{equation}
\begin{split}
T &= \begin{bmatrix} 
1 & 0 & 0 & t_x \\ 
0 & 1 & 0 & t_y \\ 
0 & 0 & 1 & t_z \\ 
0 & 0 & 0 & 1 
\end{bmatrix}, \\
R_x &= \begin{bmatrix} 
1 & 0 & 0 & 0 \\ 
0 & \cos r_x & -\sin r_x & 0 \\ 
0 & \sin r_x & \cos r_x & 0 \\ 
0 & 0 & 0 & 1 
\end{bmatrix}, \\
R_y &= \begin{bmatrix} 
\cos r_y & 0 & \sin r_y & 0 \\ 
0 & 1 & 0 & 0 \\ 
-\sin r_y & 0 & \cos r_y & 0 \\ 
0 & 0 & 0 & 1 
\end{bmatrix}, \\
R_z &= \begin{bmatrix} 
\cos r_z & -\sin r_z & 0 & 0 \\ 
\sin r_z & \cos r_z & 0 & 0 \\ 
0 & 0 & 1 & 0 \\ 
0 & 0 & 0 & 1 
\end{bmatrix}, \\
S &= \begin{bmatrix} 
s_x & 0 & 0 & 0 \\ 
0 & s_y & 0 & 0 \\ 
0 & 0 & s_z & 0 \\ 
0 & 0 & 0 & 1 
\end{bmatrix}, \\
M &= T R_z R_y R_x S
\end{split}
\end{equation}
In cases where composited primitives do not utilize the full set of transformation components—translation, rotation, and scale—we omit the respective term in the $M$ formula. For uniform scaling, represented by a scalar scaling factor $s$, we set $s_x$, $s_y$, and $s_z$ all equal to $s$.
%For composited primitives that do not use all of translation, rotation, and scale, we remove the corresponding term in the formula of $M$. We make $s_x = s_y = s_z = s$ for the uniform scale that is described by a scalar scaling $s$.

\subsection{Complete List of 3D Strokes}

\begin{table*}[t]
\centering
\begin{tabular}{l|ccccc}
\toprule
Stroke Name       & Base SDF         & Translation & Rotation & Uniform Scale & Anisotropic Scale \\
\midrule
Sphere            & Unit Sphere      &      \cmark &   \xmark &        \cmark &            \xmark \\
Ellipsoid         & Unit Sphere      &      \cmark &   \cmark &        \xmark &            \cmark \\
Axis-aligned Cube & Unit Cube        &      \cmark &   \xmark &        \cmark &            \xmark \\
Oriented Cube     & Unit Cube        &      \cmark &   \cmark &        \cmark &            \xmark \\
Axis-aligned Box  & Unit Cube        &      \cmark &   \xmark &        \xmark &            \cmark \\
Oriented Box      & Unit Cube        &      \cmark &   \cmark &        \xmark &            \cmark \\
Round Cube        & Unit RoundCube   &      \cmark &   \cmark &        \cmark &            \xmark \\
Round Box         & Unit RoundCube   &      \cmark &   \cmark &        \xmark &            \cmark \\
Line              & Unit Line        &      \cmark &   \cmark &        \cmark &            \xmark \\
Triprism          & Unit Triprism    &      \cmark &   \cmark &        \cmark &            \xmark \\
Octahedron        & Unit Octahedron  &      \cmark &   \cmark &        \cmark &            \xmark \\
Tetrahedron       & Unit Tetrahedron &      \cmark &   \cmark &        \cmark &            \xmark \\
\bottomrule
\end{tabular}
\caption{The basic shapes and transformations used in all 3D strokes.}
\label{tab:all_strokes}
\end{table*}

Utilizing various combinations of basic geometric shapes in unit space, along with transformations including translation, rotation, and scale, enables the creation of a diverse palette of 3D strokes. These strokes exhibit distinct geometric and aesthetic stylization. The full assortment of these 3D strokes is detailed in Tab.~\ref{tab:all_strokes}.

\subsection{Spline Curves}

\subsubsection{Polynomial splines}

In Sec.~\ref{sec:spline_curves} of the main paper, we use three types of different polynomial curves that are commonly used in computer graphics. Specifically, we use the quadratic Bézier, cubic Bézier, and Catmull Rom spline, respectively. We provide the concrete definition of these curves below. All points defined here are vectors in the 3D scene space.

\paragraph{Quadratic Bézier Spline.}
A quadratic Bézier spline is defined by three control points, the start point $\mathbf{P}_0$, the end point $\mathbf{P}_2$, and middle point $\mathbf{P}_1$ that is also tangent to the $\mathbf{P}_0$ and $\mathbf{P}_2$. Note that the curve only blends toward but does not pass the middle point $\mathbf{P}_1$. The parametric form is given by:
\begin{equation}
    \mathbf{C}(t;\mathbf{P}_0,\mathbf{P}_1,\mathbf{P}_2) = (1-t)^2\mathbf{P}_0 + 2(1-t)t\mathbf{P}_1 + t^2\mathbf{P}_2
\end{equation}

\paragraph{Cubic Bézier Spline.}
A cubic Bézier spline introduces an additional control point compared with the quadratic spline. This spline is defined by four points: the start point $\mathbf{P}_0$, two control points $\mathbf{P}_1$ and $\mathbf{P}_2$, and the end point $\mathbf{P}_3$. The curve starts at $\mathbf{P}_0$ and ends at $\mathbf{P}_3$, with $\mathbf{P}_1$ and $\mathbf{P}_2$ influencing its shape. The parametric equation of a cubic Bézier spline is:
\begin{equation}
\begin{split}
\mathbf{C}(t;\mathbf{P}_0,\mathbf{P}_1,\mathbf{P}_2, \mathbf{P}_3) &= (1-t)^3\mathbf{P}_0 + 3(1-t)^2t\mathbf{P}_1 \\
&+ 3(1-t)t^2\mathbf{P}_2 + t^3\mathbf{P}_3
\end{split}
\end{equation}

\paragraph{Catmull Rom Spline.}
The Catmull Rom spline is another form of cubic spline, notable for its ability to %pass through all 
interpolate its control points. This spline is defined by a series of points, with the curve passing through each of these points except the first and last. One feature of the Catmull Rom spline is that the tangent at each point is determined by the line connecting the previous and next points, ensuring a smooth transition. In our implementation, we specifically use the centripetal Catmull-Rom spline, a variant of the standard Catmull-Rom spline. This type of spline is particularly advantageous for avoiding the issue of self-intersecting loops in the curve, which are common in the uniform and chordal Catmull-Rom splines. The parametric form of the Catmull Rom spline, for a segment between $\mathbf{P}_1$ and $\mathbf{P}_2$, with $\mathbf{P}_0$ and $\mathbf{P}_3$ influencing the shape, is defined as:
\begin{equation}
\begin{split}
\mathbf{C}(t;\mathbf{P}_0,\mathbf{P}_1,\mathbf{P}_2, \mathbf{P}_3) &= \frac{1}{2} [ (2\mathbf{P}_1) \\
&+ (-\mathbf{P}_0 + \mathbf{P}_2)t \\
&+ (2\mathbf{P}_0 - 5\mathbf{P}_1 + 4\mathbf{P}_2 - \mathbf{P}_3)t^2 \\
&+ (-\mathbf{P}_0 + 3\mathbf{P}_1 - 3\mathbf{P}_2 + \mathbf{P}_3)t^3]
\end{split}
\end{equation}

\subsubsection{Nearest Point Finding}

As mentioned in Sec.~\ref{sec:spline_curves} of the main paper, we need to locate the nearest point on the spline in order to compute the SDF of the spline curve. This involves solving the following equation that finds the $t$ value with the minimized distance to the query point $\mathbf{p} \in \mathbb{R}^3$ in a differential way:
\begin{equation}
    t^* = \underset{t}{\arg\min} \Vert \mathbf{C}(t,\theta_s^\text{curve}) - \mathbf{p} \Vert_2, \; s.t. \; 0 \le t \le 1 \;,
\end{equation}
where $\theta_s^\text{curve}$ denotes the parameters of the spline curve.
An analytical solution might exist for some specific formulations of the splines. However, for more versatility, we use a general approximation solution that can be adapted to any parametric spline curve in the main paper. $K + 1$ samples are uniformly selected on the curve to form $K$ line segments, and the distance of each line segment to the query point is calculated to find $t^*$. Assuming the line segment is given as $L(t; \mathbf{A}, \mathbf{B}) = (1 - t)\mathbf{A} + t\mathbf{B}, \; t \in [0, 1]$, where $\mathbf{A}$ and $\mathbf{B}$ are the two endpoints of the line segment, the distance from point $\mathbf{p}$ to this line segment $L$ can be calculated as:
\begin{equation}
    d_L(\mathbf{p}, \mathbf{A}, \mathbf{B}) = \Vert \mathbf{p} - L(t'; \mathbf{A}, \mathbf{B}) \Vert_2
\end{equation}
where $t'$ is the $t$ value that gives the nearest point on the line segment:
\begin{equation}
    t'(\mathbf{p}, \mathbf{A}, \mathbf{B}) = \min(\max(\frac{(\mathbf{p} - \mathbf{A}) \cdot (\mathbf{B} - \mathbf{A})}{(\mathbf{B} - \mathbf{A}) \cdot (\mathbf{B} - \mathbf{A})}, 0), 1)
\end{equation}
We then compute $t^*$ using Algorithm~\ref{alg:nearest_point_finding}. The computation complexity can be easily controlled by adjusting the number of $K$, where a higher $K$ leads to a more accurate approximation but at the cost of higher computation.

\begin{algorithm}
\caption{Find the approximated nearest point on a given spline curve by uniformly sample $K$ line segments on the curve.}
\label{alg:nearest_point_finding}
\KwIn{$\mathbf{C}$: parametric spline function defining the coordinates at distance $t \in [0,1]$.}
\ExtIn{$\theta_s^{curve}$: parameters of the spline curve.}
\ExtIn{$K$: number of line segments used.}
\ExtIn{$\mathbf{p}$: coordinate of the query point.}
\KwResult{$t^*$: the $t$ value of the nearest point on the spline curve.}
$t^* \gets 0$\;
$t_\text{start} \gets 0$\;
$d_\text{min} \gets \infty$\;
$\mathbf{A} \gets \mathbf{C}(t_\text{start},\theta_s^{curve})$\;
\For{$i \gets 1$ \KwTo $K$}{
  $t_\text{end} \gets i / K$\;
  $\mathbf{B} \gets \mathbf{C}(t_\text{end},\theta_s^{curve})$\;
  \If{$d_L(\mathbf{p}, \mathbf{A}, \mathbf{B}) < d_\text{min}$}{
    $d_\text{min} = d_L(\mathbf{p}, \mathbf{A}, \mathbf{B})$\;
    $t^* = t_\text{start} + (t_\text{end} - t_\text{start})\cdot t'(\mathbf{p}, \mathbf{A}, \mathbf{B})$\;
  }
  $t_\text{start} \gets t_\text{end}$\;
  $\mathbf{A} \gets \mathbf{B}$\;
}
\end{algorithm}

\section{Implementation Details}
\label{sec:impl_details}

We implement the stroke field using Pytorch~\cite{paszke2019pytorch} framework and implement the strokes using fused CUDA kernels to accelerate training and reduce GPU memory usage. We transform sampled coordinates to the canonical volume according to the scene bounding box and use the normalized scene coordinates as inputs to the stroke field. Like ZipNeRF~\cite{barron2023zip}, we use a proposal network based on hash grid representation~\cite{muller2022instant} to facilitate ray sampling. Specifically, for each ray of a pixel, we first sample 32 points using the proposal MLP to obtain sampling weights. We resample 32 points and compute the stroke field's density and color on each point, and use the same volumetric rendering formula in NeRF to acquire the final pixel color.

We train 15k steps using 500 strokes for scenes with a single object, and train 25k steps using 1000 strokes for face-forwarding scenes. We employ the AdamW~\cite{loshchilov2017decoupled} optimizer with betas $(0.9, 0.99)$, setting the learning rate to $0.01$ and exponentially decays to $0.0003$. We start with $k_\delta = 7$ and gradually decay it to $k_\delta = 1$ during training. Additionally, as brushes are progressively added to the scene, we start with 25\% of the sampling points and gradually use all the sampling points when training reaches 80\%. We set $\lambda_{color} = 1$, $\lambda_{mask} = 0.02$, $\lambda_{den\,reg} = 0.0001$, $\lambda_{err} = 0.1$, and $\lambda_{err\,reg} = 0.001$ in our training setup.

% \section{Visualization of the error field}
% \label{sec:vis_errorfield}

\section{Quantitative comparison of strokes}
\label{sec:stroke_metrics}

In Sec.~\ref{sec:exp_recon} of the main paper, we conduct a qualitative comparison between the visual effects of several selected 3D strokes. Additionally, this section includes a comprehensive quantitative analysis of all 3D strokes, as detailed in Tab.~\ref{tab:metrics_strokes}. The metrics are measured using 500 strokes on object scenes and 1000 strokes on face-forwarding scenes.
As shown in the table, the ellipsoid stroke typically demonstrates superior fidelity in scene reconstruction, followed closely by the cubic Bézier curve.

\begin{table}[!t]
  \caption{Averaged quantitative metrics of reconstruction results of different 3D strokes.}
  \label{tab:metrics_strokes}
  \centering
  %\resizebox{0.9\columnwidth}{!}{
  \begin{tabular}{lccc}
    \toprule
    & PSNR$\uparrow$ & SSIM$\uparrow$ & LPIPS$\downarrow$ \\
    \midrule
    Sphere & $19.72$ & $0.591$ & $0.416$ \\
    Ellipsoid & $\mathbf{21.78}$ & $\mathbf{0.687}$ & $\mathbf{0.283}$ \\
    Cube & $19.65$ & $0.593$ & $0.389$ \\
    Axis-aligned Box & $19.99$ & $0.588$ & $0.408$ \\
    Oriented Box & $20.66$ & $0.637$ & $0.321$ \\
    Round Cube & $20.17$ & $0.623$ & $0.380$ \\
    Tetrahedron & $20.30$ & $0.625$ & $0.372$ \\
    Octahedron & $20.09$ & $0.614$ & $0.380$ \\
    Triprism & $20.68$ & $0.640$ & $0.351$ \\
    Line & $20.74$ & $0.641$ & $0.347$ \\
    \midrule
    Quadratic Bézier & $21.32$ & $0.676$ & $0.311$ \\
    Cubic Bézier & $21.64$ & $\mathbf{0.687}$ & $0.308$ \\
    Catmull-Rom & $21.47$ & $0.675$ & $0.324$ \\
    \bottomrule
  \end{tabular}
  %}
\end{table}

\section{Training setup of applications}
\label{sec:app_setups}

\subsection{Color Transfer}

In Sec.~\ref{sec:app_color} of the main paper, we transfer the color distribution from a reference style image to a trained stroke-based 3D scene. We adopt perceptual style loss, which extracts the gram matrix~\cite{gatys2016image} at specific layers of a pre-trained VGG16 network~\cite{johnson2016perceptual}, and computer the difference between the rendered RGB image and the target style image. Since computing style loss requires an image rather than individual pixels as input, we randomly render 32x32 chunks of original images under the given camera poses. We add this style loss with a weight $\lambda_{style}=0.25$ to the total loss and fine-tune the color parameters of trained strokes for 5k iterations.

\subsection{Text-driven scene drawing}

In Sec.~\ref{sec:app_generation} of the main paper, we use the vision-language model CLIP~\cite{radford2021learning} (ViT-B/32) to achieve scene drawing based on a given text prompt. Specifically, we minimize the loss between the text embedding of the given prompt and the image embedding of the rendered RGB image. When rendering the images, we sample camera poses in a circular path looking at the scene's origin with azimuth angle in $[0, 360]$ and elevation angle in $[40, 105]$, and render an image chunk of size $128 \times 128$. We consider the azimuth angle starting at zero as the front view and adjust the CLIP guidance scale for larger azimuth angles accordingly. This adjustment results in generation outcomes that are more coherent with the viewpoint.

For text guidance, we use the template ``\texttt{realistic 3D rendering painting of [OBJECT]}", where \texttt{[OBJECT]} is substituted with the specific object description we aim to generate. 
Additionally, we discovered that incorporating a silhouette loss alongside the RGB loss significantly enhances the quality of the generated geometric shapes. This is achieved by generating another image embedding from the rendered opacity and encouraging lower loss between this silhouette image embedding with the text embedding.

%{
%    \small
%    \bibliographystyle{ieeenat_fullname}
%    \bibliography{main}
%}

\end{document}